\documentclass[preprint,12pt,authoryear]{elsarticle}

\usepackage{amssymb}
\usepackage{amsmath}

\usepackage{multirow}
\usepackage{booktabs}
\usepackage[colorlinks,linkcolor=blue]{hyperref}
\usepackage{colortbl}
\definecolor{lightgray}{gray}{0.9}
\usepackage{algorithm}    
\usepackage{algpseudocode} 
\usepackage{amsmath}        

\usepackage{soul}

\usepackage[table]{xcolor}

\usepackage{tabularx}   
\usepackage{booktabs}    
\usepackage[table]{xcolor}
\usepackage{multirow}    

\newcommand{\CheckmarkBold}{\textbf{\checkmark}}

\journal{Nuclear Physics B}

\begin{document}

\begin{frontmatter}



\title{SAMamba: Adaptive State Space Modeling with Hierarchical Vision for Infrared Small Target Detection}


\author[1]{Wenhao Xu}
\author[1]{Shuchen Zheng} 
\author[2,3,4,5]{Changwei Wang}
\author[1]{Zherui Zhang} 
\author[6]{Chuan Ren} 
\author[3,4]{Rongtao Xu}
\author[1]{Shibiao Xu \corref{cor1}} 

\cortext[cor1]{Corresponding author}
 
\fntext[fn1]{W.Xu and S.Zheng contributed equally to this work.}
 
\address[1]{School of Artificial Intelligence, Beijing University of Posts and Telecommunications, Beijing, 100083, China.}

\address[2]{Key Laboratory of Computing Power Network and Information Security, Ministry of Education, Shandong Computer Science Center (National Supercomputer Center in Jinan), Qilu University of Technology (Shandong Academy of Sciences),Shandong, 250353, China}

\address[3]{The State Key Laboratory of Multimodal Artificial Intelligence Systems, Institute of Automation, Chinese Academy of Sciences, Beijing, 100190, China}

\address[4]{School of Artificial Intelligence, University of Chinese Academy of Sciences, Beijing, 100190, China.}

\address[5]{Shandong Provincial Key Laboratory of Computer Networks, Shandong Fundamental Research Center for Computer Science, Shandong, 250353, China}

\address[6]{
School of Software Microelectronics, Peking University, Beijing, 100871, China.
}


\begin{abstract}

Infrared small target detection (ISTD) is vital for long-range surveillance systems, particularly in military defense, maritime monitoring, and early warning applications. Despite its strategic importance, ISTD remains challenging due to two fundamental limitations: targets typically occupy less than 0.15\% of the image area and exhibit low distinguishability from complex backgrounds. While recent advances in deep learning have shown promise, existing methods struggle with information loss during downsampling and inefficient modeling of global context. This paper presents SAMamba, a novel framework that synergistically integrates SAM2's hierarchical feature learning with Mamba's selective sequence modeling to address these challenges. Our key innovations include: (1)Feature Selection Adapter (FS-Adapter) that enables efficient domain adaptation from natural to infrared imagery by employing a dual-stage selection mechanism, which includes token-level selection via a learnable task embedding and channel-wise refinement through adaptive transformations; (2)Cross-Channel State-Space Interaction (CSI) module that achieves efficient global context modeling through selective state space modeling with linear complexity; and (3)Detail-Preserving Contextual Fusion (DPCF) module that adaptively combines multi-scale features through learnable fusion strategies, utilizing a gating mechanism to balance contributions from high-resolution and low-resolution features. SAMamba effectively addresses the core challenges of ISTD by bridging the domain gap, maintaining fine-grained target details, and efficiently modeling long-range dependencies. Extensive experiments on NUAA-SIRST, IRSTD-1k and NUDT-SIRST datasets demonstrate that SAMamba significantly outperforms state-of-the-art methods, particularly in challenging scenarios with heterogeneous backgrounds and varying target scales. 
Code is available at \href{https://github.com/zhengshuchen/SAMamba}{\color{blue}https://github.com/zhengshuchen/SAMamba}.

\end{abstract}




\end{frontmatter}

\section{Introduction}
\begin{figure}[!ht]
\begin{center}
 \includegraphics[width=8cm, height=6.5cm]{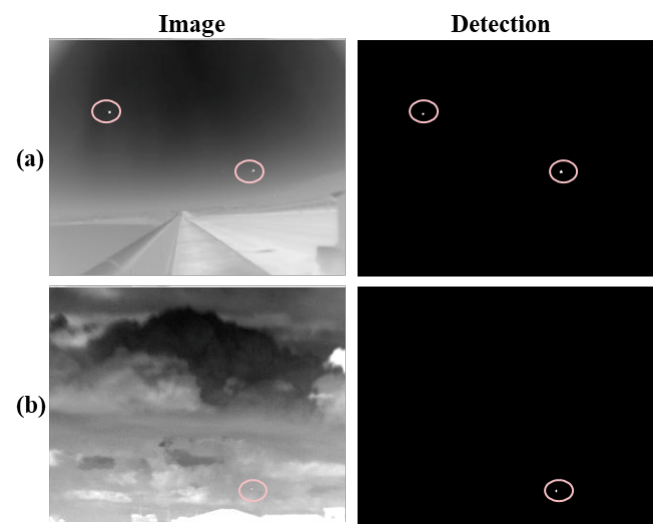}
\end{center}
   \caption{Challenges in infrared small target detection. (a) Targets are too small, making them prone to being missed during detection. (b) Targets have low distinguishability from the background.}
\label{fig:intro}
\end{figure}

Infrared small target detection (ISTD) plays a crucial role in long-range surveillance systems, particularly in military defense, maritime monitoring, and aerospace applications. Consider, for instance, the need to detect distant unmanned aerial vehicles (UAVs), monitor maritime traffic for small vessels, or identify potential aerial threats in complex environments. These applications rely on the ability to accurately detect targets that occupy a minuscule portion of the infrared sensor's field of view, often under challenging conditions of low visibility and complex backgrounds. 
 Such sensing capabilities are increasingly integrated into networked systems, where advancements in areas like deep learning are also being applied to optimize operational aspects such as energy efficiency in wireless sensor network frameworks \citep{2024using}.
However, ISTD faces two fundamental challenges arising from the inherent characteristics of infrared imaging: (1) the extreme scale discrepancy, where targets typically occupy less than 0.15\% of the image area with minimal distinguishable features (Figure~\ref{fig:intro} (a)), and (2) the homogeneous thermal signatures between targets and complex backgrounds, leading to exceptionally low signal-to-clutter ratios (Figure~\ref{fig:intro} (b)).

The development of ISTD methods has evolved through several stages. Traditional approaches include filter-based methods\citep{Zeng2006TheDO} that enhance local contrast through various filtering operations, human visual system-inspired methods\citep{chen2013local,WLCM} that mimic biological perception mechanisms, and low-rank-based algorithms~\citep{Gao2013InfraredPM,PSTNN,RIPT,dai2016infrared,NARM} that decompose images into target and background components. While these methods established important theoretical foundations, their effectiveness is largely limited to controlled environments with relatively simple backgrounds, primarily due to their reliance on hand-crafted features and restrictive assumptions about background uniformity.

Deep learning has revolutionized ISTD by introducing end-to-end trainable architectures. Early approaches focused on architectural innovations, such as adversarial frameworks for balancing missed detections and false alarms\citep{wang2019miss} and asymmetric context modulation for enhanced feature extraction\citep{dai2021asymmetric}. More advanced methods have emphasized multi-scale representation and context modeling, including the integration of local contrast priors with deep features\citep{ACLNet}, edge-aware designs with bidirectional attention\citep{ISNet}, and detail-guided feature compensation\citep{zhang2022exploring}. Recent approaches have explored dual U-Net architectures\citep{9989433}, hierarchical context fusion\citep{xu2024hcf}, and contrast-shape representation learning\citep{CSRNet}. However, these CNN-based methods face inherent limitations: multiple downsampling steps common in hierarchical architectures risk losing crucial spatial information vital for detecting diminutive targets, while the constrained local receptive field of convolutions limits their ability to effectively model the long-range dependencies necessary for distinguishing small targets from complex, spatially extensive background clutter.

The emergence of Segment Anything Model v2 (SAM2) \citep{ravi2024sam} and Vision Mamba (Vim)\citep{zhu2024vision} presents new opportunities for ISTD. SAM2 provides robust multi-scale representations through its hierarchical architecture, crucial for handling extreme scale variations. Simultaneously, Mamba implements selective state-space modeling with linear computational complexity, enabling efficient modeling of long-range dependencies critical for target-background discrimination. However, directly applying these advances presents two key challenges: (1) the significant domain gap between natural images and infrared imagery requires careful adaptation strategies, and (2) the unique characteristics of infrared small targets - sparse spatial distribution and low feature distinctiveness - demand specialized architectural considerations.

To address these limitations and leverage the potential of recent foundation models, we propose \textbf{SAMamba}, a novel framework that synergistically integrates SAM2 and Mamba for ISTD. Our approach differs significantly from existing solutions: unlike prior CNN-based methods that struggle with information loss and limited context, and distinct from naive applications of foundation models that overlook domain-specific needs and efficiency, SAMamba employs a tailored, synergistic design. Our approach is built on the insight that effective small target detection requires three complementary capabilities: (1) robust domain adaptation and task-specific feature selection, (2) efficient global context modeling for target-background discrimination, and (3) adaptive information preservation across scales. We achieve this through three key innovations: First, we introduce a \textbf{Feature Selection Adapter (FS-Adapter)} specifically designed to efficiently bridge the natural-to-infrared domain gap while selecting salient features for small targets, directly addressing the adaptation challenge. Second, we design a \textbf{Cross-Channel State-Space Interaction (CSI)} module that leverages Mamba's selective state spaces for efficient global context modeling with linear complexity, tackling the long-range dependency limitation of CNNs and the efficiency concerns of Transformers . Third, we propose a \textbf{Detail-Preserving Contextual Fusion (DPCF)} module that adaptively combines multi-scale features through learnable fusion strategies, explicitly mitigating information loss during feature aggregation. While SAMamba demonstrates significant improvements, potential limitations include the computational resources required for the large foundation model backbone and potential sensitivity to the diversity and representativeness of the training data, particularly for unseen complex background types.

Our primary contributions can be summarized as follows:
\begin{enumerate}
    \item We present a novel framework called \textbf{SAMamba} for integrating hierarchical vision and selective sequence modeling in the context of ISTD, synergistically combining SAM2's hierarchical feature learning with Mamba's adaptive processing to address fundamental challenges in small target detection.
    \item We propose the \textbf{FS-Adapter} module that enables efficient domain adaptation while preserving target-specific features through a dual-stage selection mechanism, bridging the gap between natural and infrared imagery.
    \item We design the \textbf{CSI} module that leverages Mamba's selective state spaces to achieve efficient and focused global context modeling, particularly suited for the sparse and low-contrast nature of infrared targets.
    \item We introduce the \textbf{DPCF} module that adaptively combines multi-scale features through learnable fusion strategies, significantly improving the preservation and integration of target-relevant information.
    \item Through extensive experiments on NUAA-SIRST, IRSTD-1k, and NUDT-SIRST datasets, we demonstrate that SAMamba achieves significant improvements over state-of-the-art methods.
\end{enumerate}

The remainder of this paper is organized as follows: Section \ref{sec:related} provides a review of related work on infrared small target detection and relevant vision foundation models. Section \ref{sec:method} details the proposed SAMamba framework, including the FS-Adapter, CSI, and DPCF modules. Section \ref{sec:exp} presents the experimental setup, evaluation metrics, main results, and ablation studies on benchmark datasets. Finally, Section \ref{sec:conclusion} concludes the paper and summarizes our findings.

\section{Related Work}
\label{sec:related}
\subsection{Infrared small target detection (ISTD)}
The evolution of ISTD methods can be traced through two major paradigms: traditional model-based approaches and deep learning methods.

\noindent\textbf{Traditional Methods.} primarily rely on hand-crafted features and prior assumptions. Filter-based approaches~\citep{Zeng2006TheDO} enhance target saliency through local contrast manipulation but struggle with complex backgrounds. Human visual system-inspired methods~\citep{chen2013local,WLCM} attempt to mimic biological perception mechanisms by incorporating directional sensitivity and weighted local contrast, showing improved robustness in scenarios with distinguishable targets. Low-rank-based methods~\citep{Gao2013InfraredPM,PSTNN,RIPT,dai2016infrared,NARM} decompose images into target and background components through matrix/tensor factorization, achieving better performance in varying backgrounds. However, these methods often make strong assumptions about background uniformity and target characteristics, limiting their applicability in real-world scenarios.

\noindent\textbf{Deep Learning Methods.} have significantly advanced ISTD performance through learned feature representations. Early approaches focused on improving feature extraction through architectural innovations, such as adversarial learning for balancing missed detections and false alarms~\citep{wang2019miss}, and asymmetric context modulation~\citep{dai2021asymmetric} for enhanced feature discrimination. Recent works have explored the integration of traditional priors with deep features, exemplified by local contrast fusion~\citep{ACLNet} and edge-aware feature extraction~\citep{ISNet}. Multi-scale feature learning has emerged as a key direction, with methods like detail-guided feature compensation~\citep{zhang2022exploring}, resolution-preserving supervision~\citep{9989433}, and hierarchical context fusion~\citep{xu2024hcf} addressing the scale variation challenge. Despite these advances, CNN-based approaches still face fundamental limitations in preserving small target information and capturing long-range dependencies.

\subsection{Vision Foundation Models for Dense Prediction}
Recent vision foundation models have demonstrated remarkable potential for dense prediction tasks, with SAM and Mamba representing two significant advances in architectural design.

\noindent\textbf{SAM and Its Variants:} The Segment Anything Model (SAM)~\citep{sam} introduced a powerful ViT-based encoder that has been widely adapted across vision tasks. Key developments include adapter-based fine-tuning for high-frequency feature extraction~\citep{SAM-Adapter,gao2024ts}, temporal-spatial consistency learning~\citep{hui2024endow}, and domain-specific adaptations~\citep{luo2024sam,cheng2024unleashing}. The recent SAM2~\citep{ravi2024sam}, featuring the Hiera~\citep{ryali2023hiera} hierarchical encoder, offers enhanced multi-resolution feature extraction capabilities. Notable applications have demonstrated its effectiveness in medical imaging~\citep{chen2024sam2} and multi-task scenarios~\citep{xiong2024sam2}. These advances suggest the potential of hierarchical feature learning for small target detection, though domain adaptation remains challenging.

\noindent\textbf{Mamba-based Architectures:} The Mamba architecture~\citep{gu2023mamba} represents a paradigm shift in sequence modeling through selective state-space models (SSMs). Its linear computational complexity and superior long-range dependency modeling have sparked numerous visual computing innovations. Vision Mamba (ViM)~\citep{zhu2024vision} adapted SSMs for high-resolution image processing through bidirectional modeling and positional embeddings. Subsequent works have explored various architectural integrations, including U-Net combinations for medical imaging~\citep{ma2024u,liu2024swin}, multi-directional scanning for underwater enhancement~\citep{an2024uwmamba}, and cross-modal fusion for object detection~\citep{wang2024mask}. Of particular relevance to ISTD is the impact of channel-wise processing on model efficiency~\citep{wu2024ultralight} and the potential for global context modeling in high-resolution imagery~\citep{zhao2024rs}.

\section{Method}
\label{sec:method}
\subsection{Preliminaries}
\noindent \textbf{Segment Anything Model v2:} 
SAM2 advances vision foundation models through its hierarchical architecture, primarily through its Hiera backbone that enables systematic multi-scale feature extraction. Given an input image $\mathbf{I}\in\mathbb{R}^{3\times H\times W}$, Hiera generates a sequence of hierarchical features:
\begin{equation}
\mathbf{X}_i\in\mathbb{R}^{C_i\times \frac{H}{2^{i+1}}\times \frac{W}{2^{i+1}}}, \quad i\in{1,2,3,4}
\end{equation}
where, channel dimensions follow $C_i\in{96,192,384,768}$ for the small variant. Unlike traditional vision transformers that maintain uniform resolution, this hierarchical design systematically reduces spatial dimensions while increasing feature channels, enabling efficient multi-scale representation particularly beneficial for small target detection.

\noindent \textbf{Vision Mamba:}
Vision Mamba (ViM) extends the selective state space modeling paradigm to visual tasks through a bidirectional scanning mechanism. Given an input feature map $\mathbf{F}\in\mathbb{R}^{H\times W\times C}$, ViM processes it through state space transformations:
\begin{equation}
\mathbf{h}_t = \overline{A}\mathbf{h}_{t-1} + \overline{B}\mathbf{x}_t, \quad \mathbf{y}_t = \mathbf{C}\mathbf{h}_t,
\end{equation}
where, $\overline{A}\in\mathbb{R}^{N\times N}$, $\overline{B}\in\mathbb{R}^{N\times 1}$, and $\mathbf{C}\in\mathbb{R}^{1\times N}$ are learned parameters.\par 
The model computes selective global convolution through a structured kernel:
\begin{equation}
\mathbf{\overline{K}} = (\mathbf{C}\overline{B}, \mathbf{C}\overline{AB}, \mathbf{C}\overline{A}^2\overline{B}, ..., \mathbf{C}\overline{A}^{M-1}\overline{B}),
\end{equation}
This formulation enables efficient sequence modeling with linear computational complexity while maintaining the ability to capture long-range dependencies. The output is computed through convolution:

\begin{equation}
\mathbf{y} = \mathbf{x} * \mathbf{\overline{K}},
\end{equation}
where, $M$ is the length of the input sequence $\mathbf{x}$, and $\mathbf{\overline{K}}\in\mathbb{R}^{M}$ is a structured convolutional kernel. This architecture proves particularly effective for ISTD by enabling efficient global context modeling without the quadratic complexity of traditional attention mechanisms, crucial for discriminating small targets from complex backgrounds.

\begin{figure*}[!t]
\begin{center}
\vspace{-0.5em}    
\setlength{\abovecaptionskip}{0cm} 
\setlength{\belowcaptionskip}{0cm} 
 \includegraphics[width=13cm, height=7cm]{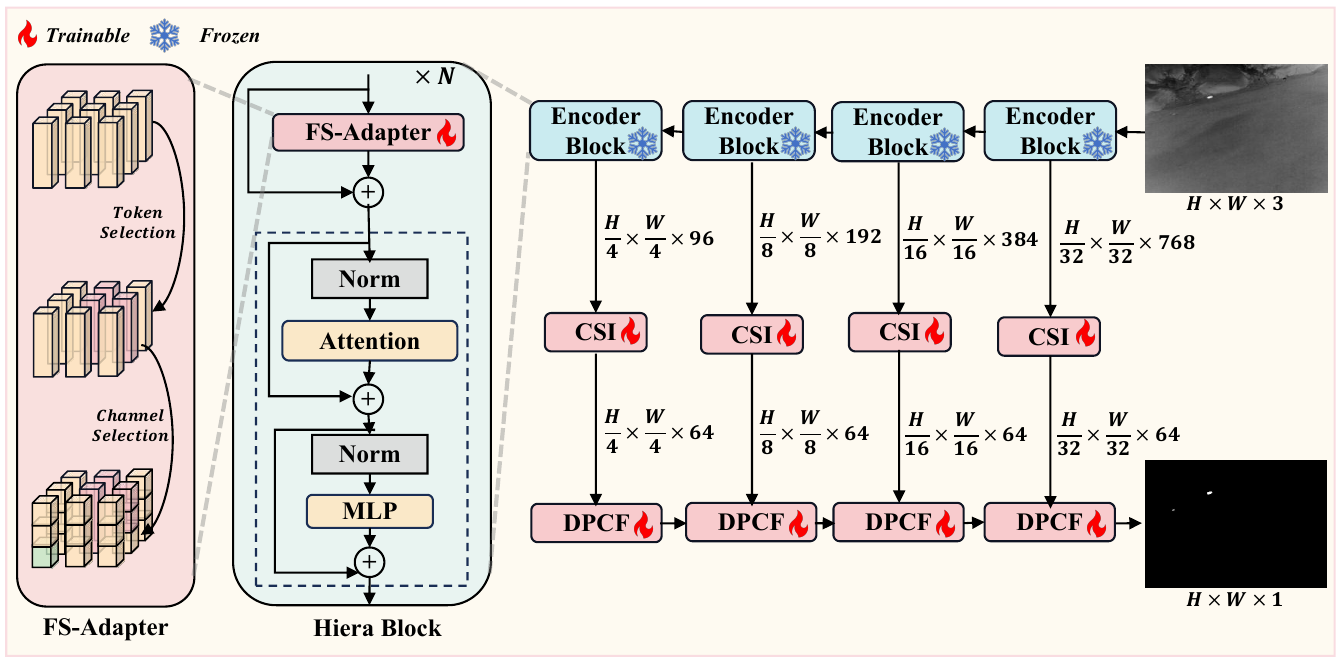}
\end{center}
  \caption{Overview of the proposed SAMamba framework. The architecture consists of three key components: Feature Selection Adapter (FS-Adapter) for domain-specific feature extraction,Cross-Channel State-Space Interaction (CSI) for long-range dependency modeling, and Detail-Preserving Contextual Fusion (DPCF) for multi-scale feature aggregation.}
\label{fig:domain_net}
\end{figure*}

\subsection{Overview}
This section outlines the overall workflow of the proposed SAMamba network, illustrating how its components synergistically process infrared images for small target detection within a U-Net architectural framework (Figure \ref{fig:domain_net}). The process begins with the input image entering the encoder, which leverages the hierarchical SAM2 Hiera backbone. Crucially, before each Hiera block, features are processed by our lightweight Feature Selection Adapter (FS-Adapter) module, enabling parameter-efficient domain adaptation and focusing the powerful hierarchical feature extraction on ISTD-relevant characteristics. As the encoder progressively downsamples the feature maps, generating multi-scale representations, features from each scale are simultaneously passed through skip connections. Within these connections, the Cross-Channel State-Space Interaction (CSI) module is employed. CSI utilizes Vision Mamba's selective state-space modeling to efficiently capture long-range spatial dependencies and enhance the global context awareness of the encoder features before they reach the decoder. The decoder then progressively upsamples feature maps from the deepest layer. At each upsampling stage, the features are fused with the corresponding context-enhanced skip connection features processed by CSI. This fusion is mediated by the Detail-Preserving Contextual Fusion (DPCF) module, which adaptively balances the contribution of fine-grained details (from skip connections) and semantic context (from upsampled features) to mitigate information loss, particularly for small targets. Finally, the output feature map from the last decoder stage is passed through a segmentation head (typically convolutional layers followed by an activation function) to produce the final pixel-wise prediction mask identifying the infrared small targets. This integrated pipeline effectively combines hierarchical feature learning (adapted by FS-Adapter in the encoder), efficient global context modeling (via CSI in skip connections), and detail-preserving fusion (by DPCF in the decoder) to address the core challenges of ISTD.

\begin{figure*}[!ht]
\setlength{\abovecaptionskip}{0cm} 
\setlength{\belowcaptionskip}{0cm} 
\begin{center}
 \includegraphics[width=13cm, height=7.5cm]{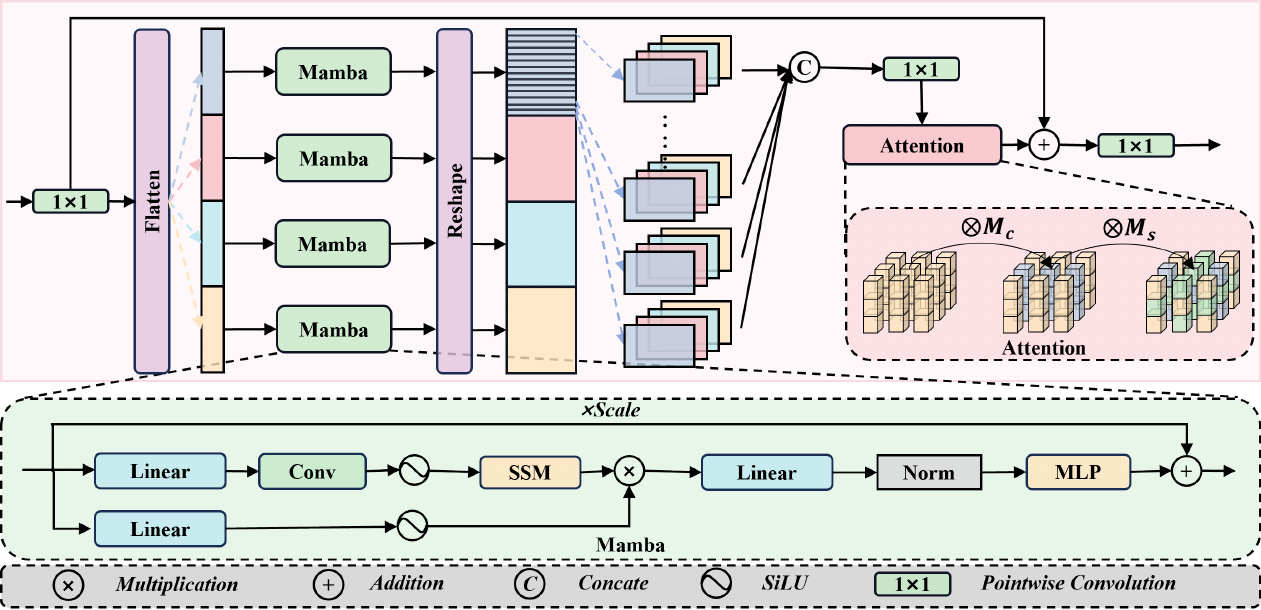}
\end{center}

\caption{Architecture of the Cross-Channel State-Space Interaction (CSI) module. The module integrates Vision Mamba blocks for efficient sequence modeling with cross-channel feature recombination and dual-attention refinement for enhanced target-background discrimination.}

\label{fig:CSI}
\end{figure*}
\subsection{Feature Selection Adapter (FS-Adapter)}

While SAM2 provides powerful hierarchical representations, directly applying its pre-trained features to ISTD poses two challenges: (1) the domain gap between natural images and infrared imagery, and (2) the need for task-specific feature emphasis on small targets. To address these challenges while maintaining computational efficiency, we propose the Feature Selection Adapter (FS-Adapter), a lightweight module inserted before each Hiera block in the encoder. It enables targeted adaptation of pre-trained features through dynamic token and channel selection, focusing the model on ISTD-relevant information.
Given the relatively limited size of typical ISTD datasets, we leverage the pre-trained variant, SAM2 Hiera-S, and adopt a parameter-efficient fine-tuning strategy. Specifically, we freeze the pre-trained Hiera parameters and integrate FS-Adapter modules before each block. The encoder generates four feature maps at distinct scales:
\begin{equation}
    (\mathbf{F}_i)_{i=1}^{4}\in\mathbb{R}^{\frac{H}{2^{i+1}}\times\frac{W}{2^{i+1}}\times C_i}
\end{equation}
For Hiera-S, the channel dimensions are defined as: $C_i\in\{96, 192, 384, 768\}$.

The FS-Adapter performs adaptive feature refinement through two key mechanisms: token-level selection and channel-wise refinement. 
For token selection, given an input feature $\mathbf{F}_{t}\in\mathbb{R}^{H\times W\times C}$ reshaped to $\mathbf{F}_{t}'\in\mathbb{R}^{HW\times C}$, we compute:
\begin{equation}
    \hat{\mathbf{t}}_i = \mathbf{t}_i \cdot \text{sim}(\mathbf{t}_i,\mathbf{\xi}),
    \label{eq:token_sel}
\end{equation}
where, $(\hat{\mathbf{t}}_i)_{i=1}^{C} \in \mathbb{R}^{HW}$ represents the $i$-th token of the selected tokens $\hat{\mathbf{F}}_t$, $(\mathbf{t}_{i})_{i=1}^{C} \in \mathbb{R}^{HW}$ represents the $i$-th token of $\mathbf{F}_{t}'$, and $\mathbf{\xi} \in \mathbb{R}^{C}$ is a learnable task embedding that encodes ISTD-specific feature importance. The similarity function $\text{sim}(\cdot, \cdot)$ is implemented as normalized cosine similarity, ensuring stable training:
\begin{equation}
    \text{sim}(\mathbf{a},\mathbf{b}) = \max(0,\frac{\mathbf{a}^T\mathbf{b}}{\|\mathbf{a}\|\|\mathbf{b}\|})
\label{eq:sim}
\end{equation}
This step effectively re-weights each channel's spatial map ($\mathbf{t}_i$) based on its relevance to the ISTD task as encoded in $\mathbf{\xi}$. The resulting weighted channel maps $\hat{\mathbf{t}}_i$ are reassembled into $\hat{\mathbf{F}}_t' \in \mathbb{R}^{HW \times C}$.

The weighted tokens $\hat{\mathbf{F}}_t'$ undergo channel-wise refinement. This involves a linear transformation using a learnable matrix $\mathbf{P} \in \mathbb{R}^{C \times C}$, which allows interaction and mixing of information across channels. The result is reshaped back to the spatial dimensions and added to the original input feature via a residual connection:
\begin{equation}
\mathbf{F}_o = \text{Conv}(\text{Reshape}(\hat{\mathbf{F}}_t' \mathbf{P})) + \mathbf{F}_{t}, \label{eq:channel_ref}
\end{equation}

where $\text{Conv}$ represents a 1x1 convolution, and $\text{Reshape}$ operations handle the transitions between sequence and spatial formats. The residual connection $\mathbf{F}_{t}$ is crucial for preserving the knowledge learned during pre-training while allowing the adapter to make task-specific adjustments. This dual-stage selection mechanism effectively adapts SAM2 features for ISTD with minimal trainable parameters.

\subsection{Cross-Channel State-Space Interaction (CSI)}
Inspired by the recent success of state-space models~\citep{gu2023mamba}, we adapt the Vision Mamba (ViM)\citep{zhu2024vision} architecture within our CSI module, placed in the skip connections, to effectively capture long-range dependencies while maintaining linear computational complexity. As shown in Figure~\ref{fig:CSI}, the CSI module performs the following steps:

The input feature map from the encoder skip connection undergoes a $1 \times 1$ convolution for channel dimension alignment with the corresponding decoder stage. It is then spatially flattened into a sequence $\mathbf{F}_{m}\in\mathbb{R}^{HW\times C}$. Recognizing the impact of input channel count on Mamba's parameter growth~\citep{wu2024ultralight}, we employ a channel-wise parallel processing strategy. The sequence $\mathbf{F}_{m}$ is divided into four segments along the channel dimension: $(\mathbf{m}'_{i})_{i=1}^{4}\in\mathbb{R}^{HW\times\frac{C}{4}}$.

Each segment $\mathbf{m}'_{i}$ is processed independently by a dedicated Vision Mamba (VIM) block. This block consists of the core Mamba layer for sequence modeling, followed by Layer Normalization (LN) and a Multi-Layer Perceptron (MLP). A scaled residual connection, controlled by a learnable scaling factor $\gamma$, is added to facilitate information flow and stable training. The output of each VIM block is $\mathbf{m}_i$, reshaped back to spatial dimensions $\mathbb{R}^{H\times W\times\frac{C}{4}}$. This parallel processing mitigates parameter explosion and enhances computational efficiency.
\begin{equation}
\mathbf{m}_i=MLP(LN(Mamba(\mathbf{m}'_{i})))+\gamma\mathbf{m}'_{i},
\end{equation}
where $LN$ denotes Layer Normalization, and $\gamma$ is a learnable scalar scaling factor.

 To enhance feature complementarity and allow interaction between the parallel streams, we introduce a cross-channel segmentation and recombination scheme (Figure~\ref{fig:CSI}). Each output feature map $\mathbf{m}_i$ is further segmented channel-wise into single-channel maps $(\mathbf{m}_i^j)_{j=1}^{C/4}\in\mathbb{R}^{H\times W\times 1}$. Channels corresponding to the same channel index $j$ but originating from different Mamba heads $i$ are then concatenated together, forming $C/4$ feature groups $(\mathbf{h}_j)_{j=1}^{C/4}\in\mathbb{R}^{H\times W\times 4}$.
\begin{equation}
\mathbf{h}_j =[\mathbf{m}_1^j, \mathbf{m}_2^j, \mathbf{m}_3^j, \mathbf{m}_4^j] \quad \text{for } j=1, ..., C/4
\end{equation}
These recombined groups are then concatenated back along the channel dimension: $[\mathbf{h}_1, \mathbf{h}_2, ..., \mathbf{h}_{C/4}] \in \mathbb{R}^{H\times W\times C}$.

A pointwise ($1 \times 1$) convolution, denoted $W_{outer}$ (representing its learnable weights), is applied to the fully recombined feature map. This lightweight step effectively fuses information across the recombined channels, followed by Batch Normalization ($\mathcal{B}$) and SiLU activation ($\delta$). This promotes a holistic representation and refines features, focusing on relevant information.
\begin{equation}
\mathbf{F}o =\delta(\mathcal{B}(W_{outer}([\mathbf{h}_1, \mathbf{h}_2, ..., \mathbf{h}_{C/4}]))),
\end{equation}
where $W_{outer}$ applies the $1 \times 1$ convolution.

Finally, we utilize sequential channel and spatial attention mechanisms (similar in principle to CBAM~\citep{woo2018cbam} but potentially with simpler implementations) to adaptively adjust feature importance. The output $\mathbf{F}_o \in \mathbb{R}^{H \times W \times C}$ is first multiplied element-wise by a channel attention map $\mathbf{M}_c \in \mathbb{R}^{1 \times 1 \times C}$ (computed from global pooling of $\mathbf{F}_o$), and the result is then multiplied by a spatial attention map $\mathbf{M}_s \in \mathbb{R}^{H \times W \times 1}$ (computed via convolutions on pooled channel information). This highlights target-relevant channels and spatial locations while suppressing background noise.
\begin{equation}
\begin{aligned}
\mathbf{F}_c &= \mathbf{M}_c \odot \mathbf{F}_o,  \\
\mathbf{F}_s &= \mathbf{M}_s \odot \mathbf{F}_c, 
\end{aligned}
\end{equation}
where $\mathbf{F}_s \in \mathbb{R}^{H \times W \times C}$ is the final output of the CSI module, passed to the decoder.
\begin{figure*}[t]
\setlength{\abovecaptionskip}{0cm}
\setlength{\belowcaptionskip}{0cm}
\begin{center}
\includegraphics[width=13cm, height=5cm]{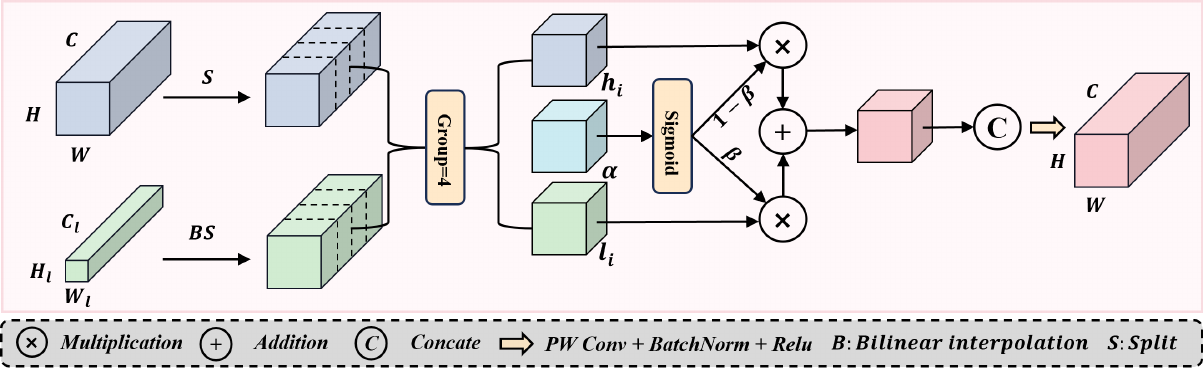}
\end{center}
\caption{Illustration of the Detail-Preserving Contextual Fusion (DPCF) module. Low-res features ($\mathbf{F}_l$) are upsampled (interpolated) to match high-res features ($\mathbf{F}_h$). Both are split channel-wise into four segments. A learnable parameter $\alpha$ generates spatial gating weights $\beta$ via sigmoid. These gates control the weighted sum of corresponding high-res ($\mathbf{h}_i$) and low-res ($\mathbf{l}_i$) segments. The fused segments ($\mathbf{o}'_i$) are concatenated and refined by a final convolution block.}
\label{fig:dpcf}
\end{figure*}

\subsection{Detail-Preserving Contextual Fusion (DPCF)}
A critical challenge in ISTD is the potential for small target information in high-resolution feature maps to be diluted or lost when fused with lower-resolution, context-rich feature maps during decoder upsampling. To counteract this, we propose the Detail-Preserving Contextual Fusion (DPCF) module, illustrated in Figure~\ref{fig:dpcf}, used at each stage of the decoder where skip connection features are fused with upsampled deeper features. DPCF dynamically integrates features from these different scales using a learnable, spatially-adaptive gating mechanism.
The process within DPCF is as follows:

The low-resolution feature map from the deeper decoder layer, $\mathbf{F}_l\in\mathbb{R}^{H_l\times W_l\times C_l}$, is first upsampled using bilinear interpolation to match the spatial dimensions $(H, W)$ of the high-resolution feature map $\mathbf{F}_h\in\mathbb{R}^{H\times W\times C}$ (typically coming from the CSI module via skip connection). Channels are also aligned, usually via $1 \times 1$ convolutions if needed, resulting in both maps having dimension $\mathbb{R}^{H\times W\times C}$. Both aligned feature maps are then partitioned along the channel dimension into four equal segments: $(\mathbf{h}_i)_{i=1}^{4}\in\mathbb{R}^{H\times W\times\frac{C}{4}}$ and $(\mathbf{l}_i)_{i=1}^{4}\in\mathbb{R}^{H\times W\times\frac{C}{4}}$, representing segments of the high-res and low-res features, respectively.

A single learnable parameter $\alpha' \in\mathbb{R}^{1\times 1\times 1}$ is introduced. This parameter is expanded spatially and channel-wise to match the shape of the feature segments, resulting in $\alpha \in\mathbb{R}^{H\times W\times\frac{C}{4}}$. This expanded parameter is passed through a sigmoid function to generate gating weights $\beta \in [0, 1]$ for each spatial position and channel segment.
\begin{equation}
\beta = \text{sigmoid}(\alpha), \quad \beta \in\mathbb{R}^{H\times W\times\frac{C}{4}}
\end{equation}
These weights $\beta$ control the contribution of the low-resolution features, while $(1-\beta)$ controls the contribution of the high-resolution features for each corresponding segment $i$.

 The adaptive fusion is performed via a weighted sum for each segment:
\begin{equation}
\mathbf{o}_i' = \beta \odot \mathbf{l}_i + (1-\beta) \odot \mathbf{h}_i,
\end{equation}
where $\odot$ denotes element-wise multiplication, and $\mathbf{o}_i'\in\mathbb{R}^{H\times W\times\frac{C}{4}}$ is the selectively aggregated feature map for the $i$-th segment. This allows the network to learn, for each spatial location and channel group, whether to prioritize fine details from $\mathbf{h}_i$ (when $\beta$ is close to 0) or contextual information from $\mathbf{l}_i$ (when $\beta$ is close to 1).

The adaptively fused segments $(\mathbf{o}_i')_{i=1}^{4}$ are concatenated back along the channel dimension to form $\mathbf{F}'_o\in\mathbb{R}^{H\times W\times C}$. Finally, a convolutional block, typically consisting of a $3 \times 3$ convolution ($Conv(\cdot)$), followed by Batch Normalization ($\mathcal{B}(\cdot)$) and a ReLU or SiLU activation ($\delta(\cdot)$), refines the fused feature map, yielding the final output $\mathbf{F}_o\in\mathbb{R}^{H\times W\times C}$ for that decoder stage.
\begin{equation}
\begin{aligned}
\mathbf{F}_o' &= [\mathbf{o}_1', \mathbf{o}_2', \mathbf{o}_3', \mathbf{o}_4'], \\
\mathbf{F}_o &= \delta(\mathcal{B}(\text{Conv}(\mathbf{F}_o'))),
\end{aligned}
\end{equation}
This mechanism empowers the network to selectively emphasize channels and spatial regions within both high- and low-dimensional features that are most discriminative for small targets, fostering a nuanced, detail-aware fusion with minimal computational overhead.

\subsection{Loss Function}
We apply Deconvolution and Convolution to the output of the final DPCF module to generate a predicted mask, which is compared with the ground truth for loss computation. The loss function consists of SoftIoU Loss\citep{rahman2016optimizing}, Dice Loss\citep{milletari2016v} and Focal Loss\citep{ross2017focal}:
\begin{equation}
 \mathcal{L} = SoftIoU(y, \hat{y}) + Dice(y, \hat{y}) + Focal(y, \hat{y}), 
\end{equation}
where, \(\hat{y}\) represents the ground truth, \(y\) is the predicted mask, and \(\mathcal{L}\) is the final loss.

\begin{figure*}[htbp]
\begin{center}
 \includegraphics[width=14cm, height=12cm]{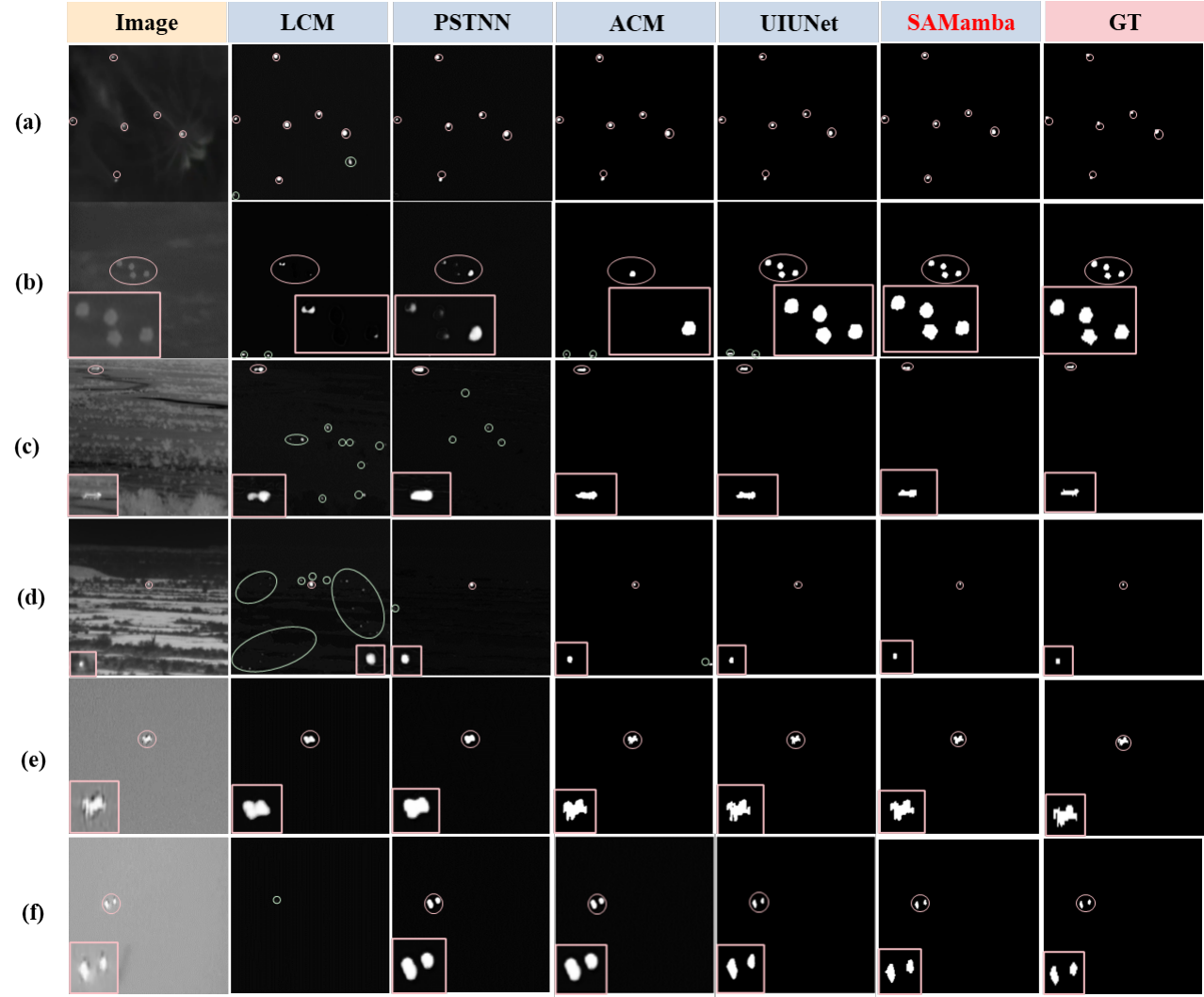}
\end{center}
   \caption{Visual examples of representative methods are provided. Pink and green circles represent true-positive and false-positive objects, respectively. Objects marked within the pink rectangles are zoomed in for a clearer comparison of detection accuracy among different methods.
   }
\label{fig:visual}
\end{figure*}

\section{Experiments}
\label{sec:exp}
\subsection{Experimental Setup}
We conduct extensive experiments to validate the effectiveness of SAMamba on three challenging ISTD benchmarks: NUAA-SIRST~\citep{dai2021asymmetric}, IRSTD-1k~\citep{zhang2022isnet}, and NUDT-SIRST~\citep{li2022dense}.
\begin{itemize}
\item The NUAA-SIRST dataset contains 427 real infrared images where targets typically occupy less than 0.1\% of the image area, featuring complex backgrounds with significant clutter.
\item The IRSTD-1k dataset comprises 1,001 real images from diverse operational scenarios including maritime, urban, and natural environments, presenting additional challenges through varied background textures and noise patterns.
\item The NUDT-SIRST dataset consists of 1327 high-quality synthetic images covering five typical background types  and various target types. A significant portion (96\%) are small targets (0.15\% area), with 27\% being extremely small (0.01\% area), and many exhibiting low brightness, simulating real-world low signal-to-noise ratio challenges.
\end{itemize}

SAMamba is implemented in PyTorch and trained on an NVIDIA RTX 3090 GPU. All input images are processed at 1024×1024 resolution through dynamic scaling and cropping operations. The network is optimized using Adam~\citep{adam} with an initial learning rate of 1e-4, which is reduced by a factor of 0.1 every 100 epochs. Training proceeds for 300 epochs with a batch size of 2. For the Hiera backbone, we initialize with pre-trained SAM2-S weights and employ our proposed FS-Adapter for domain-specific fine-tuning.

\subsection{Evaluation Metrics}
We formulate ISTD as a semantic segmentation task and evaluate performance using three complementary metrics: Intersection over Union (IoU), normalized Intersection over Union (nIoU), and F1-score.

The IoU metric, widely adopted in semantic segmentation, measures the overlap between predicted ($P$) and ground truth ($T$) target regions at the pixel level, aggregated over the dataset:
\begin{equation}
\begin{aligned}
IoU &= \frac{A_{inter}}{A_{union}}
&= \frac{\sum_{i=1}^{N} TP_i}{\sum_{i=1}^{N}(T_i + P_i - TP_i)},
\end{aligned}
\label{eq:iou}
\end{equation}
where $N$ is the number of test samples, $TP_i$ is the number of true positive pixels for sample $i$, $T_i$ is the number of ground truth target pixels, and $P_i$ is the number of predicted target pixels.

The nIoU metric~\citep{dai2021asymmetric} extends IoU by averaging the IoU score calculated for each sample individually, which can better reflect performance on datasets with varying target sizes and numbers per image:
\begin{equation}
\begin{aligned}
nIoU=\frac{1}{N}\sum_{i=1}^{N}\frac{TP_i}{T_i+P_i-TP_i},
\end{aligned}
\label{eq:niou}
\end{equation}
where the terms are defined per-sample as above.

The F1-score is the harmonic mean of Precision (P) and Recall (R), providing a balanced measure, particularly useful in cases of class imbalance common in ISTD:
\begin{equation}
\begin{aligned}
F_1 &= \frac{1}{N}\sum_{i=1}^{N} 2 \cdot \frac{Precision_i \cdot Recall_i}{Precision_i + Recall_i}, 
\end{aligned}
\label{eq:f1}
\end{equation}
Here, $\text{Precision}_i = \frac{TP_i}{P_i},\ \text{Recall}_i = \frac{TP_i}{T_i}$.
 Precision and Recall are typically calculated per image and then averaged.

\subsection{Main Results}

\begin{table*}[!htbp]
\centering
\caption{Comprehensive comparison of ISTD methods on three datasets. Performance measured by IoU(\%), nIoU(\%), and F1-Score(\%).}
\label{tab:total_results_combined}
\renewcommand{\arraystretch}{1.2} 
\setlength{\tabcolsep}{1pt} 
\begin{tabular}{@{}l *{9}{c}@{}}
\toprule
\multirow{2}{*}{\textbf{Method}} & 
\multicolumn{3}{c}{\textbf{NUAA-SIRST}} & 
\multicolumn{3}{c}{\textbf{IRSTD-1k}} & 
\multicolumn{3}{c}{\textbf{NUDT-SIRST}} \\
\cmidrule(lr){2-4} \cmidrule(lr){5-7} \cmidrule(lr){8-10}
& IoU & nIoU & $F_1$ & IoU & nIoU & $F_1$ & IoU & nIoU & $F_1$ \\
\midrule
\rowcolor{gray!10} 
\multicolumn{10}{@{}l}{\textit{\textbf{Traditional Methods}}} \\
Top-Hat$_{\textit{IPT}'2006}$ & 5.86 & 25.42 & 14.63 & 4.26 & 15.08 & 16.02 & 20.72 & 28.98 & 33.52 \\
LCM$_{\textit{TGRS}'2013}$ & 6.84 & 8.96 & -- & 4.45 & 4.73 & -- & -- & -- & -- \\
WLCM$_{\textit{GRSL}'2018}$ & 22.28 & 28.62 & -- & 9.77 & 16.07 & -- & -- & -- & -- \\
NARM$_{\textit{RS}'2018}$ & 25.95 & 32.23 & -- & 7.77 & 12.24 & -- & 6.93 & 6.19 & -- \\
PSTNN$_{\textit{RS}'2019}$ & 39.44 & 47.72 & 68.00 & 16.44 & 25.91 & 27.48 & 27.72 & 39.80 & 43.41 \\
IPI$_{\textit{TIP}'2013}$ & 40.48 & 50.95 & 73.12 & 14.40 & 16.12 & 26.05 & 37.49 & 48.38 & 54.53 \\
RIPT$_{\textit{JSTARS}'2017}$ & 25.49 & 33.01 & 44.21 & 8.15 & 16.12 & 20.35 & 29.17 & 36.12 & 45.16 \\
NIPPS$_{\textit{IPT}'2016}$ & 33.16 & 40.91 & -- & 16.38 & 27.10 & -- & -- & -- & -- \\
\midrule
\rowcolor{gray!10} 
\multicolumn{10}{@{}l}{\textit{\textbf{Deep Learning Methods}}} \\
MDvsFA$_{\textit{ICCV}'2019}$ & 56.17 & 59.84 & -- & 50.85 & 45.97 & -- & 75.14 & 73.85 & -- \\
ACM$_{\textit{WACV}'2021}$ & 72.45 & 72.15 & 80.87 & 63.38 & 60.80 & 77.59 & 68.48 & 69.26 & 81.29 \\
ALCNet$_{\textit{TGRS}'2021}$ & 74.31 & 73.12 & 82.92 & 62.05 & 59.58 & 75.47 & 81.40 & 80.71 & 78.59 \\
ISNet$_{\textit{CVPR}'2022}$ & 80.02 & 78.12 & 81.49 & 68.77 & 64.84 & 81.49 & 84.94 & 84.13 & 85.23 \\
FC3-Net$_{\textit{MM}'2022}$ & 74.22 & 72.46 & -- & 64.98 & 63.59 & -- & -- & -- & -- \\
UIUNet$_{\textit{TIP}'2023}$ & 78.25 & 75.15 & 86.95 & 62.66 & 61.35 & 79.63 & 88.91 & 89.60 & 94.23 \\
CSRNet$_{\textit{TIP}'2024}$ & 77.93 & 78.72 & -- & 65.87 & 66.70 & -- & -- & -- & -- \\
HCFNet$_{\textit{ICME}'2024}$ & 80.09 & 78.31 & 88.95 & 63.44 & 62.84 & 77.34 & 75.70 & 74.97 & 80.42 \\
\rowcolor{yellow!20} 
\textbf{Ours} & \textbf{81.08} & \textbf{79.17} & \textbf{89.55} & \textbf{73.53} & \textbf{68.99} & \textbf{84.75} & \textbf{93.13} & \textbf{93.15} & \textbf{96.44} \\
\bottomrule
\end{tabular}
\footnotetext{Higher values indicate better performance. Best results are in bold. '-' indicates unavailable values.}
\end{table*}

As observed in Table~\ref{tab:total_results_combined}, traditional methods, relying on hand-crafted features and prior assumptions, exhibit limited performance across all datasets. Their IoU scores generally remain below 41\%, nIoU below 51\%, and F1 scores below 74\%, particularly struggling with the complex backgrounds and target variations present in IRSTD-1k and NUDT-SIRST. This underscores the difficulty these methods face in adapting to diverse and challenging real-world and synthetic scenarios.
Deep learning methods demonstrate substantial improvements across the board, showcasing the power of learned representations. Recent architectures like ISNet, UIUNet, and HCFNet achieve competitive results, often exceeding 75\% IoU/nIoU and 80\% $F_1$ on NUAA-SIRST and showing strong performance on the other datasets. For instance, ISNet leverages edge awareness, HCFNet employs hierarchical context fusion, and UIUNet uses a dual U-Net structure, each contributing to advancements in the field.
Crucially, our proposed SAMamba consistently achieves state-of-the-art performance across all three datasets and all three evaluation metrics. On NUAA-SIRST, SAMamba reaches 81.08\% IoU, 79.17\% nIoU, and 89.55\% $F_1$. On the more diverse IRSTD-1k, it achieves 73.53\% IoU, 68.99\% nIoU, and 84.75\% $F_1$, showing a significant margin over previous methods. 
Notably, on the challenging synthetic NUDT-SIRST dataset with extremely small and dim targets, SAMamba demonstrates remarkable robustness, achieving 93.13\% IoU, 93.15\% nIoU, and an impressive 96.44\% F1-score. This consistent superiority across datasets with different characteristics (real vs. synthetic, varying complexity) and diverse metrics highlights the effectiveness and generalizability of SAMamba's design, which synergistically combines hierarchical feature extraction (SAM2), efficient long-range dependency modeling (Mamba via CSI), domain adaptation (FS-Adapter), and detail-preserving fusion (DPCF).

Visual comparisons in Figure~\ref{fig:visual} further validate these quantitative results. Compared to previous methods, SAMamba exhibits three key strengths: (1) improved target completeness with reduced false positives in sparse scenes (Figure~\ref{fig:visual} (a) and (b)), (2) robust discrimination against complex background clutter (Figure~\ref{fig:visual} (c) and (d)), and (3) enhanced preservation of fine-grained target characteristics (Figure~\ref{fig:visual} (e) and (f)). These capabilities directly address core ISTD challenges of scale variation and low target-background contrast.

\subsection{Ablation study}
\label{sec:ab}

To validate our architectural choices and understand the contribution and sensitivity of each component, we conduct systematic ablation studies and hyperparameter analyses on the NUAA-SIRST dataset.

\noindent\textbf{Component Contribution.} Table~\ref{tab:ablation_study} demonstrates the progressive performance improvement achieved by incrementally adding each key component to a baseline U-Net architecture. Starting from the baseline (71.20\% IoU), incorporating the SAM2 Hiera-S encoder provides a substantial +4.23\% IoU gain (reaching 75.43\% IoU), confirming the benefit of its hierarchical features. Adding the FS-Adapter for parameter-efficient domain adaptation further improves performance by +0.89\% IoU (reaching 76.32\% IoU). Introducing the CSI module in the skip connections yields another significant gain of +2.47\% IoU (reaching 78.79\% IoU), highlighting the importance of its enhanced global context modeling. Finally, integrating the DPCF module for adaptive feature fusion in the decoder results in a final boost of +2.28\% IoU, reaching our full model's performance of 81.08\% IoU and underscoring the effectiveness of preserving detail during fusion.

\begin{table}[!htbp]
  \centering
  \caption{Component Ablation Analysis of SAMamba.  
          The baseline (Bas.) is built upon U-Net architecture. 
          Each component's contribution is evaluated using IoU(\%) and nIoU(\%) metrics on the NUAA-SIRST dataset.  
          \textbf{Checkmarks (\CheckmarkBold) indicate the presence of each component.}
          Hiera$^*$: SAM2's hierarchical encoder.}
  \label{tab:ablation_study} 
 
  \setlength{\tabcolsep}{6pt} 
  \renewcommand{\arraystretch}{1.2} 
 
  \begin{tabular}{@{}ccccc|cc@{}}
    \toprule[1.2pt] 
    \textbf{Base} & \textbf{Hiera$^*$} & \textbf{FS-Adapter} & \textbf{CSI} & \textbf{DPCF} 
    & \multicolumn{1}{c}{\textbf{IoU (\%)}} & \multicolumn{1}{c@{}}{\textbf{nIoU (\%)}} \\
    \midrule
    \CheckmarkBold &  &  &  &  & 71.20 & 74.40 \\ 
    \CheckmarkBold & \CheckmarkBold &  &  &  & 75.43 & 75.20 \\
    \CheckmarkBold & \CheckmarkBold & \CheckmarkBold &  &  & 76.32 & 75.45 \\[1pt] %
    \CheckmarkBold & \CheckmarkBold & \CheckmarkBold & \CheckmarkBold &  & 78.79 & 78.35 \\
    \rowcolor{yellow!20} 
    \CheckmarkBold & \CheckmarkBold & \CheckmarkBold & \CheckmarkBold & \CheckmarkBold 
    & \textbf{81.08} & \textbf{79.17} \\
    \bottomrule[1.2pt] 
  \end{tabular}

\end{table}

\begin{figure*}[]
\setlength{\abovecaptionskip}{0cm} 
\setlength{\belowcaptionskip}{0cm} 
\begin{center}
 \includegraphics[width=13.5cm, height=4cm]{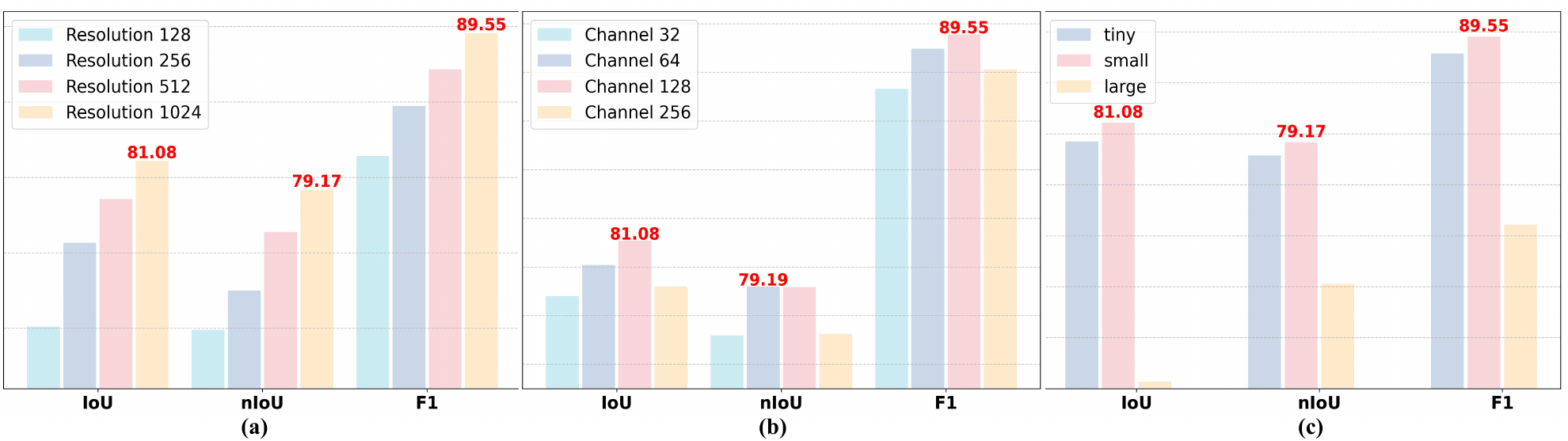}
\end{center}
   \caption{Impact of key architectural parameters on SAMamba's performance. (a) Model performance across different input resolutions, showing the trade-off between spatial detail preservation and computational cost. (b) Analysis of CSI module channel dimensions, demonstrating the relationship between feature capacity and model efficiency. (c) Comparison of different Hiera backbone variants (Tiny, Small, Large), illustrating the impact of model capacity on detection performance. Results are reported on NUAA-SIRST dataset using both IoU and nIoU metrics.}
\label{fig:ablation}
\end{figure*}

\noindent\textbf{Architectural Parameter Study.} Parameter studies (Figure~\ref{fig:ablation}) reveal critical design insights regarding overall architecture choices. Resolution analysis (Fig~\ref{fig:ablation}(a)) shows performance improving with resolution up to 1024×1024, after which gains plateau while computational cost increases significantly. The analysis of the CSI module's  channel dimension (Fig~\ref{fig:ablation}(b)) indicates that 128 channels achieve the optimal balance between feature capacity and efficiency for this task. Comparing different Hiera backbone variants (Fig~\ref{fig:ablation}(c)) demonstrates that Hiera-S provides the best performance; Hiera-T is slightly less capable, while Hiera-L leads to performance degradation, likely due to overfitting on the relatively small ISTD datasets despite using PEFT.

\begin{table}[!htbp]
\centering
\caption{Hyperparameter Sensitivity Analysis on NUAA-SIRST Dataset.  
         Performance is measured using IoU(\%) and nIoU(\%). 
         \textbf{Default configurations used in SAMamba are highlighted.}}
\label{tab:sensitivity}
\setlength{\tabcolsep}{4pt} 
\renewcommand{\arraystretch}{1.0} 
\begin{tabular}{@{}l ccc@{}}
\toprule
\multirow{1}{*}{\textbf{Parameter Type}} & \multirow{1}{*}{\textbf{Configuration}} 
& \multicolumn{1}{c}{\textbf{IoU (\%) }} 
& \multicolumn{1}{c@{}}{\textbf{nIoU (\%)}} \\
\midrule
\multicolumn{4}{@{}l}{\textit{\textbf{(a) Sensitivity to CSI Heads (Total Channels = 128):}}} \\
& 1  & 79.05 & 78.11 \\
& 2 & 79.78 & 78.85 \\
\rowcolor{yellow!20} 
& \textbf{4} & \textbf{81.08} & \textbf{79.17}  \\
& 8 & 79.53 & 78.60 \\
\midrule
\multicolumn{4}{@{}l}{\textit{\textbf{(b) Sensitivity to DPCF Segments:}}} \\
& 1 & 79.35 & 78.44 \\
& 2  & 79.91 & 79.03 \\
\rowcolor{yellow!20} 
& \textbf{4} & \textbf{81.08} & \textbf{79.17}  \\
& 8  & 80.02 & 79.11 \\
\midrule
\multicolumn{4}{@{}l}{\textit{\textbf{(c) Comparison of DPCF Fusion Strategies:}}} \\
& Addition + Conv & 78.61 & 77.75 \\
& Concatenation + Conv & 79.12 & 78.23 \\
\rowcolor{yellow!20} 
& \textbf{Adaptive} & \textbf{81.08} & \textbf{79.17} \\
\bottomrule
\end{tabular}
\end{table}

\noindent\textbf{Hyperparameter Sensitivity Analysis.} To further assess the robustness of our design choices within the CSI and DPCF modules, we analyze the model's sensitivity to key internal hyperparameters. The results are presented in Table \ref{tab:sensitivity}.
First, we investigate the impact of the number of parallel Mamba heads used in the CSI module, keeping the total channel dimension fixed at 128. As shown in Table~\ref{tab:sensitivity} (a), using 4 heads (our default configuration) yields the best performance (81.08\% IoU). Performance decreases slightly with fewer heads (1 or 2), potentially due to reduced capacity for parallel processing of diverse channel information, and also degrades slightly with more heads (8), possibly due to fragmentation or increased optimization difficulty.
Second, we examine the sensitivity to the number of channel segments used in the DPCF module for adaptive fusion, as shown in Table \ref{tab:sensitivity} (b). The results indicate that performance is relatively stable for 2, 4, and 8 segments, with 4 segments providing the marginal best result. Using only 1 segment (equivalent to applying the same fusion weight across all channels) results in a noticeable drop in performance (79.35\% IoU), confirming the benefit of channel-segmented adaptive fusion.
Third, we compare our adaptive DPCF fusion strategy against simpler, fixed fusion methods: element-wise addition of the high-resolution and upsampled low-resolution features, and concatenation followed by a $3 \times 3$ convolution. The adaptive DPCF significantly outperforms both fixed strategies, demonstrating the value of learning context-aware fusion weights to dynamically balance detail preservation and context integration as shown in Table \ref{tab:sensitivity} (c).
Overall, these analyses indicate that while performance is somewhat sensitive to these internal hyperparameters, our chosen configurations (4 heads for CSI, 4 segments for DPCF, adaptive fusion) are well-supported by the empirical results and provide a robust design.

\subsection{Computational Analysis}
\label{sec:comp}
To provide a comprehensive understanding of the resource requirements of SAMamba, we analyze its computational complexity, model size, and runtime performance. We report the number of trainable parameters, FLOPs calculated for a single 1024x1024 input image, and the inference speed measured in Frames Per Second (FPS) on an NVIDIA RTX 3090 GPU with a batch size of 1. Table \ref{tab:computational} compares SAMamba with the baseline U-Net used in our ablation study and other representative deep learning methods.

\begin{table}[!htbp]
\centering
\caption{Computational Analysis of SAMamba and Other Methods.  
  Parameters are in Millions (M), FLOPs are in GigaFLOPs (G) for a $1024\times1024$ input, and FPS is measured on an NVIDIA RTX 3090 (Batch Size=1). 
  }
\label{tab:computational}
\setlength{\tabcolsep}{3pt} 
\renewcommand{\arraystretch}{1.4} 
\begin{tabular}{@{}l ccc@{}} 
\toprule
\textbf{Method} & \multicolumn{1}{c}{\textbf{Params (M) }} 
              & \multicolumn{1}{c}{\textbf{FLOPs (G) }} 
              & \multicolumn{1}{c@{}}{\textbf{FPS }} \\
\midrule
\multicolumn{4}{@{}l}{\textit{\textbf{Existing Methods:}}} \\
U-Net~\citep{ronneberger2015u} & 31.04 & 875.80 & 11.86 \\
ACM~\citep{dai2021asymmetric} & 0.52 & 8.06 & 44.49 \\
ISNet~\citep{ISNet}$^\dagger$ & 0.97 & 490.10 & 7.14 \\
FC3-Net~\citep{zhang2022exploring} & 6.90 & 42.29 & 35.6 \\
UIUNet~\citep{wu2022uiu} & 50.54 & 872.01 & 7.56 \\
HCFNet~\citep{xu2024hcf} & 14.40 & 372.45 & 7.46 \\
\midrule %
\multicolumn{4}{@{}l}{\textit{\textbf{SAMamba Variants:}}} \\
\rowcolor{blue!8} 
\multicolumn{4}{@{}l}{\textit{\quad\bfseries --- Hiera-T Series ---}} \\
SAMamba (CSI, $c$=32) & 28.65 & 129.73 & 7.32 \\
SAMamba (CSI, $c$=64) & 28.84 & 196.14 & 7.08 \\ 
SAMamba (CSI, $c$=128) & 29.49 & 460.47 & 6.64 \\
\midrule 
\rowcolor{green!8} 
\multicolumn{4}{@{}l}{\textit{\quad\bfseries --- Hiera-S Series ---}} \\
SAMamba (CSI, $c$=32) & 36.34 & 163.09 & 6.82 \\
SAMamba (CSI, $c$=64) & 36.53 & 229.49 & 6.61 \\
SAMamba (CSI, $c$=128) & 37.18 & 493.82 & 6.39 \\
\bottomrule
\end{tabular}
\end{table}

As shown in Table~\ref{tab:computational}, SAMamba variants demonstrate an efficient computational profile despite their sophisticated architecture. The Hiera-T series variants maintain parameter counts (28.65-29.49M) comparable to the baseline U-Net (31.04M), while the Hiera-S series exhibits a moderate increase (36.34-37.18M). Notably, all SAMamba variants achieve substantially reduced computational loads (129.73-493.82 GFLOPs) compared to the baseline U-Net (875.80 GFLOPs). This efficiency stems from the integration of parameter-efficient FS-Adapters and the linear complexity of Mamba blocks within the CSI module.

The inference speed of SAMamba variants (6.39-7.32 FPS on an RTX 3090) positions them competitively among specialized infrared small target segmentation models like ISNet (7.14 FPS) and HCFNet (7.46 FPS), though lighter architectures such as ACM (44.49 FPS) and FC3-Net (35.6 FPS) offer faster processing at the cost of reduced performance. This analysis demonstrates that SAMamba achieves state-of-the-art accuracy while maintaining reasonable computational requirements, effectively balancing performance gains with resource efficiency. The strategic combination of the Hiera backbone with selective Mamba integration proves to be both powerful and computationally feasible for practical applications.

\begin{figure}[!ht]
\setlength{\abovecaptionskip}{0cm} 
\setlength{\belowcaptionskip}{0cm} 
\begin{center}
 \includegraphics[width=11cm, height=10.5cm]{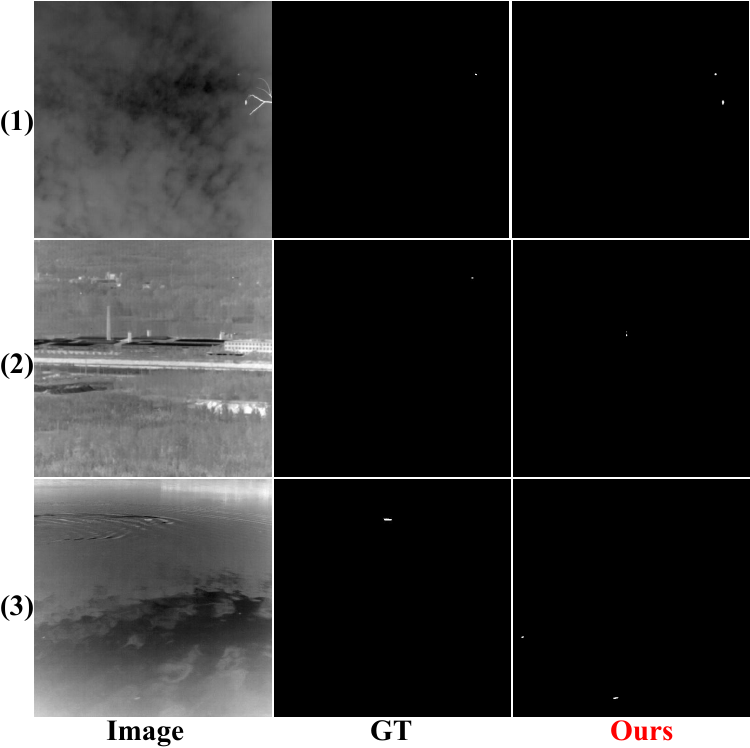}
\end{center}
\caption{Visualization of challenging scenarios illustrating potential limitations.} 

\label{fig:limit}
\end{figure}

\section{Limitations}
Despite its strong performance, SAMamba faces certain limitations inherent to challenging ISTD scenarios. Qualitative analysis reveals difficulties in two primary areas:

\noindent \textbf{Highly Complex Backgrounds.}
Targets embedded within intricate textures (e.g., dense cloud edges, complex ground clutter) can exhibit reduced feature discriminability, potentially leading to missed detections even with domain adaptation and global context modeling (Figure~\ref{fig:limit} (1) and (2)). Residual domain gap effects or fundamentally similar local statistics between target and clutter remain challenging.

\noindent \textbf{Extremely Low signal-to-clutter ratios (SCR).}
Targets with exceptionally low signal-to-clutter ratios or those significantly blended with uniform backgrounds approach fundamental detection thresholds (Figure~\ref{fig:limit} (3)). Inherent signal ambiguity can lead to non-detections, particularly under strict false alarm constraints.

\section{Conclusion}
\label{sec:conclusion}
We presented SAMamba, a novel framework for infrared small target detection that harmonizes SAM2's hierarchical vision modeling with Mamba's selective state-space sequence processing. Through the integrated FS-Adapter, CSI, and DPCF modules, SAMamba adeptly addresses the dual challenges of extreme scale discrepancy and low signal-to-clutter ratio, while preserving computational efficiency. Our extensive evaluations on the NUAA-SIRST, IRSTD-1k, and NUDT-SIRST benchmarks demonstrate that SAMamba achi-eves significant performance improvements over existing methods, establishing its effectiveness for demanding perception tasks.
Future work will focus on enhancing SAMamba's capabilities. Key directions include: leveraging temporal information from video data to improve robustness via motion analysis; optimizing the model architecture for efficient deployment on resource-limited hardware through compression techniques; and exploring multi-modal sensor fusion to achieve superior detection performance under a wider range of environmental conditions.

\bibliographystyle{elsarticle-harv} 
\bibliography{cas-refs}

\end{document}